\documentclass{article}

\usepackage[preprint, nonatbib]{neurips_2024}

\usepackage[utf8]{inputenc} % allow utf-8 input
\usepackage[T1]{fontenc}    % use 8-bit T1 fonts
\usepackage{hyperref}       % hyperlinks
\usepackage{url}            % simple URL typesetting
\usepackage{booktabs}       % professional-quality tables
\usepackage{amsfonts}       % blackboard math symbols
\usepackage{nicefrac}       % compact symbols for 1/2, etc.
\usepackage{microtype}      % microtypography
\usepackage{xcolor}         % colors
\usepackage{graphicx} 
\usepackage{multicol}
\usepackage{multirow}
\usepackage[numbers]{natbib}

\usepackage{amsmath,amssymb,amsfonts}
\usepackage[ruled]{algorithm2e}
\usepackage{amsthm}

\title{You Only Accept Samples Once: Fast, Self-Correcting Stochastic Variational Inference}

\author{%
  Dominic B.~Dayta \\
  Mathematical Informatics\\
  Nara Institute of Science and Technology\\
  \texttt{dominic.dayta.da4@naist.ac.jp} \\
}

\begin{document}

\maketitle

\begin{abstract}
We introduce YOASOVI, an algorithm for performing fast, self-correcting stochastic optimization for Variational Inference (VI) on large Bayesian heirarchical models. To accomplish this, we take advantage of available information on the objective function used for stochastic VI at each iteration and replace regular Monte Carlo sampling with acceptance sampling. Rather than spend computational resources drawing and evaluating over a large sample for the gradient, we draw only one sample and accept it with probability proportional to the expected improvement in the objective. The following paper develops two versions of the algorithm: the first one based on a naive intuition, and another building up the algorithm as a Metropolis-type scheme. Empirical results based on simulations and benchmark datasets for multivariate Gaussian mixture models show that YOASOVI consistently converges faster (in clock time) and within better optimal neighborhoods than both regularized Monte Carlo and Quasi-Monte Carlo VI algorithms.
\end{abstract}

\section{Introduction}

Stochastic Variational Inference \cite{Ranganath-2014, Paisley-2012} provides a promising, general-purpose method for approximating any arbitrary intractable distribution with a tractable function whose parameters can be searched through direct optimization of an objective known as the Evidence Lower Bound (ELBO). However, its success depends on controlling the variance of its update steps, specifically the variance of its ELBO gradient estimator. To this end, the approach has been to draw Monte Carlo samples of size $S$ of this gradient estimator, and using the average as the estimate, resulting in a variance that decays with rate $O(S^{-1})$ \cite{Buccholz-2018}.

This causes a crucial bottleneck in applying the algorithm: too high variance means the algorithm can fail to converge towards some optimal value, but too many samples means the algorithm slows down significantly. Not only is the gradient estimate evaluated over a sample of $N$ observations in the data, it must be replicated and averaged out across the $S$ Monte Carlo samples as well. In certain objective functions, particularly in large Bayesian models involving numerous global and local parameters, the required size $S$ can be very large. Recent developments have thus focused on controlling this baseline variance either through re-parametrization \cite{Xu-2019} or regularization \cite{Dayta-2024}. A very attractive alternative that has been pursued recently is to reduce the required number of samples by forcing the sample itself to be well-behaved. For instance, drawing Quasi-Monte Carlo samples \cite{Buccholz-2018, Leobacher-2014} is able to find optimal values even when samples of only size 10 are drawn.

In the present paper, we take this idea to the extreme by requiring \textit{only a single Monte Carlo sample per iteration}, and applying an accept-reject scheme depending on the quality of the sample obtained. Thus, our sample is effectively self-correcting, in the sense of acceptance sampling being self-correcting \cite{Gelman-2013}, as it has a built-in mechanism of guaranteeing that each new sample obtained meets some desired characteristic with some probability. Combined with an adaptive sampling scheme that become stricter as iterations are spent, we produce an algorithm that not only guarantees convergence, but directly ensures it ends up within some optimal neighborhood of the objective. We show that this design has a natural convergence criteria that is much easier and more intuitive to train.

Our algorithm thus achieves optimal behavior while still being general, and can easily be implemented into standard software packages similar with how the No-U-Turn Sampler (NUTS) has been packaged with Stan \cite{Carpenter-2017}. Moreover, the computational efficiency of the method guarantees that it should be feasible to use even on local machines.

The rest of this paper is thus structured as follows: Section \ref{sec:related} reviews the Monte-Carlo Variational Inference (MCVI) algorithm, along with recent theoretical findings and improvements, specifically Quasi-Monte Carlo Variational Inference (QMCVI), while Section \ref{sec:yoasovi} develops the algorithm first in its naive construction and through a formalization making use of the Metropolis algorithm. This section also highlights some important considerations regarding hyperparameter settings. Finally, Section \ref{sec:experiments} demonstrates its performance against both MCVI and QMCVI in numerical experiments.

\section{Related Work}
\label{sec:related}

\subsection{Monte Carlo Variational Inference}

Monte Carlo Variational Inference (MCVI) making use of the score function gradient estimator \cite{Ranganath-2014, Paisley-2012} marks a significant turn in the development of stochastic VI methods. The algorithm provides a general gradient estimator that can be used alongside stochastic optimization to perform VI on any arbitrary combination of conjugate or non-conjugate models, requiring from the user no more of the usual derivations that are associated with previous flavors of VI \cite{Blei-2017}. The attention it has received, both in terms of theoretical developments \cite{Domke-2020, Kim-2023, Diao-2023, Lambert-2023} and algorithmic improvements \cite{Domke-2020, Welandawe-2022} only further support its growing importance.

We present a basic development of the algorithm here and discuss improvements and additional results that have been obtained since. In line with most applications of VI, the posterior distribution $p(\theta | y)$ for a model indexed by parameters $\theta$ on observed values $y$ are approximated by a tractable distribution $q(\theta | \lambda)$ indexed by a free parameter $\lambda$. The parameter $\lambda$ is set such that the resulting approximation $q$ is as close as possible to the posterior $p$. This can be achieved by maximizing an objective function known as the Evidence Lower Bound (ELBO) given by
\begin{align}
    \label{eq:ELBO}
    \mathcal{L}(\lambda) = \mathbb{E}[\log p(y,\theta) - \log q(\theta | \lambda)]
\end{align}
whose gradient with respect to the variational parameters $\lambda$ can be expressed by a similar expectation
\begin{align}
    \label{eq:ELBOgrad}
    \nabla_{\lambda} \mathcal{L}(\lambda) = \mathbb{E}[\nabla_{\lambda} \log q(\theta | \lambda)(\log p(y,\theta) - \log q(\theta | \lambda))]
\end{align}
The function $\nabla_{\lambda} \log q(\theta | \lambda)$ is known in classical statistics as the \textit{score function}, hence making (\ref{eq:ELBOgrad}) the score function ELBO gradient. As the goal of VI is to maximize Equation (\ref{eq:ELBO}), we can perform stochastic optimization of the form
\begin{align}
    \label{eq:stochOpt}
    \lambda_t = \lambda_{t - 1} + \rho_t \hat\nabla_{\lambda} \mathcal{L}(\lambda_{t-1})
\end{align}
where $\rho_t$ is some learning rate, and $\hat\nabla_{\lambda} \mathcal{L}(\lambda_{t-1})$ is a Monte Carlo estimate of the gradient based on the last-updated values of the parameter $\lambda_{t-1}$, obtained by drawing a reasonably large number of Monte Carlo draws $\theta[s] \sim q(\theta | \lambda)$, $s = 1,2,...,S$ and obtaining the average
\begin{align}
\label{eq:stochVI}
    \hat\nabla_{\lambda} \mathcal{L}(\lambda) = \frac{1}{S} \sum^{S}_{s = 1} \nabla_{\lambda} \log q(\theta[s] | \lambda)(\log p(y,\theta[s]) - \log q(\theta[s] | \lambda))
\end{align}

The success of MCVI is highly dependent on the variance of this score gradient estimator, and for this reason the standard approach is to set $S$ to some large value (usually $S = 500$ at the minimum, or as much as $S = 1,000 - 10,000$). \citet{Kingma-2013} introduce an alternative reparametrization gradient for MCVI, under which the variance has been shown to be much lower \cite{Xu-2019}, but at present the applicability of this behavior is restricted to specific variational families with a differentiable mapping. Rao-Blackwellization \cite{Casella-1996}, which is used by \citet{Ranganath-2014}, solves this variance problem by requiring a prior step that modifies the score gradient estimator per parameter $\lambda$ that appears in the model. Both effectively result in algorithms that are less general than the original proposal for MCVI, although they do control the behavior of Naive MCVI.

Other modifications have attempted to retain the generality of the algorithm while also controlling its variance. Regularization methods, specifically through an application of the James-Stein estimator \cite{Dayta-2024} in classical statistics yielding a heuristic that is similar to gradient clipping in used in deep neural network training \cite{Koloskova-2023,Goodfellow-2016}, promises some success, although another branch takes advantage of importance sampling \cite{Ruiz-2016}. A more comprehensive review of recent findings together with some unifying perspective can be gleaned from the work by \citet{Zhang-2018}.

\subsection{Quasi-Monte Carlo Variational Inference}

A simple yet effective improvement over standard MCVI is to make use of much more evenly distributed random samples through Quasi-Monte Carlo (QMC) methods. Although suggested towards the end of the paper by \citet{Ranganath-2014}, and subsequently used for a specific application by \citet{Tran-2017}, this approach is not extensively explored and benchmarked until \citet{Buccholz-2018}. QMCVI exhibit some desired properties \cite{Leobacher-2014} in that QMC generally exhibit low discrepancy, and are much more evenly distributed along their support compared to regular MC samples. This improved uniformity means that samples are much better behaved even with low sample sizes, leading to faster convergence brought by requiring smaller number of computations.

Implementing QMCVI simply requires replacing the sampling scheme when drawing $\theta[s]$. First, a deterministic uniform sequence is drawn for instance using either a scrambled Sobol sequence \cite{Antonov-1979} or a Halton sequence \cite{Halton-1964}. Then, this sequence is transformed to replicate the target distribution. To draw from a normal distribution $\mathcal{N}(\mu, \Sigma)$ from a Sobol sequence $u$, we simply apply the transformation $z = \Phi^{-1}(u)\Sigma^{1/2} + \mu$ for $\Sigma^{1/2}$ the upper-triangular Cholesky decomposition of the covariance matrix $\Sigma$. Drawing the Sobol sequence can be done automatically with the \texttt{randtoolbox} \cite{Christophe-2023} package in \texttt{R}. Afterwards, the rest of the MCVI algorithm applies.

Empirical results presented by \citet{Buccholz-2018} use only samples of size $S = 10$ and yet are able to reach higher values of the ELBO within faster clock times. In comparison, MCVI with similar sample sizes generally do not yield meaningful results (although this can potentially be improved with some form of regularization), and the only comparable performance for MCVI is when using samples of size $S = 100$. However, while the resulting samples are definitely much better behaved, there remains a tendency for the algorithm to make U-turns and effectively re-start from a less optimal position than before.

We nevertheless obtain our motivation for the rest of the paper. Not only is MCVI inefficient due to the uneven samples resulting from MC sampling compared to QMC, but the larger required sample size $S$ means that operations are repeated a large number of times per iteration. Driving down the sample size per iteration should be possible so long as the sample is well-behaved. In our proposed algorithm, we argue that only one sample is needed, so long as this single sample yields a consistent improvement over the previous ELBO. Not only does this further shorten computation time, but also removes the tendency for the algorithm to make U-turns in the middle of its run.

\section{You Only Accept Samples Once}
\label{sec:yoasovi}

We now discuss our proposed algorithm, YOASOVI (You Only Accept Samples Once for VI). Specifically, we produce two versions, with the former being a naive intuition that directly constructs our desired behavior into the algorithm. The second version is a more formal approach, building up YOASOVI as a Metropolis algorithm applied on MCVI, but we show that this formalization is neatly approximated by our naive intution.

Central to the development of this paper is the argument that with better-behaved samples, there is no need to expend computational resources towards large sample sizes $S$. At each iteration, a sample of size one ought to be sufficient to push the optimizer towards better and better values of the objective function. We begin by noting that embedded into the Monte Carlo score gradient used in both MCVI and QMCVI (\ref{eq:stochVI}) is the expression needed to also estimate the ELBO via a similar averaging,
\begin{align}
\label{eq:stochELBO}
    \hat{\mathcal{L}}(\lambda) = \frac{1}{S} \sum^{S}_{s = 1} (\log p(y,\theta[s]) - \log q(\theta[s] | \lambda))
\end{align}

This highlights an interesting property of Stochastic VI among the class of stochastic optimization problems, in that at each iteration $t$, after drawing a set of samples for the model parameter $\theta[s]$, we obtain with only trivial calculations both the update step $\hat\nabla_{\lambda} \mathcal{L}(\lambda)$ and the resulting objective $\hat{\mathcal{L}}(\lambda)$. No additional sampling or cycles through the data are required to evaluate both values.

What this means is that at each iteration, we can obtain the proposed update step and the expected quality of this update step in terms of whether this improves the previous objective value already obtained. Suppose that at the start of a certain iteration $t$, the previous iteration's objective is given by $\mathcal{L}_{t-1}$. After drawing the Monte Carlo (or Quasi-Monte Carlo) samples, the proposed update step is then computed to result in a new ELBO $\mathcal{L}_t$. We can then propose a \textit{self-correcting} \cite{Gelman-2013} algorithm that accepts this new iteration only if it results in an improvement over $\mathcal{L}_{t-1}$. Otherwise, the algorithm skips this iteration and waits until it comes across an update that does make such an improvement.

We take this argument a step further. We can limit the number of Monte Carlo draws $S$ down to as small as $S = 1$. In other words, with a well-behaved, self-correcting sample, there should be no need to enlarge $S$, apply regularization, or use quasi-random number generation to control the resulting sample variance. Instead, at each iteration, we need to \textit{only accept a sample once}. We now intuit a naive formulation of YOASOVI as follows. We propose a sample $\hat\nabla_{\lambda} \mathcal{L}(\lambda)$ by drawing a single value for $\theta \sim q(\theta | \lambda)$. The resulting value $\mathcal{L}_t$ then represents the new ELBO associated with this proposed sample.

The update represented by $\hat\nabla_{\lambda} \mathcal{L}(\lambda)$ is acceptable if its resulting $\mathcal{L}_t$ is better than $\mathcal{L}_{t-1}$ by some amount. If $\mathcal{L}_t \geq \mathcal{L}_{t-1}$, then we can accept this sample with certainty as a definite improvement. Otherwise, we rapidly tail off the probability of acceptance based on how far $\mathcal{L}_{t}$ deviates away from $\mathcal{L}_{t-1}$. How rapid this tailing off is done can be controlled by introducing a slope hyperparameter $M$.

More succinctly, we can accept the proposed sample with probability
\begin{align}
    p_{accept} = \min\bigg(1, 1 + \frac{M (\mathcal{L}_{t} - \mathcal{L}_{t-1})}{\mathcal{L}_{t-1}} \bigg)
\end{align}
for some constant $M > 0$. For now we defer our discussion for the optimal value(s) of $M$ and say only that it can be set to some well-behaved constant or following some adaptive logic. Regardless, if the decision is to accept the sample, then we continue with the usual updating step for stochastic optimization. Otherwise, no updating is done, and a new sample of the gradient is drawn and subjected to the same accept/reject scheme.

\begin{algorithm}[ht]
    \label{alg:naive}
    \caption{YOASOVI with naive acceptance sampling}

    \DontPrintSemicolon
    \SetAlgoLined
    \SetKwInOut{Input}{Input}\SetKwInOut{Output}{Output}
    \Input{Model, patience parameter, learning rate $\rho_t$}
    Initialize $\lambda_0$ randomly, set $t=0$, $\nu_t = 0$, and $\mathcal{L}_0 = -\infty$\;
    \BlankLine
    \While{$\nu_t < \text{patience}$}{
      $t = t+1$\;
      // Draw a single MC sample from $q(\theta | \lambda_{t-1})$ \;
      $\theta[1] \sim q(\theta | \lambda_{t-1})$ \;
      $u \sim \text{Uniform}(0,1)$ \;
      $\hat\nabla_\lambda \mathcal{L}(\lambda_{t-1}) = \nabla_{\lambda} \log q(\theta[1] | \lambda_{t-1}) (\log p(y, \theta[1]) - \log q(\theta[1] | \lambda_{t-1}))$ \;
      $\mathcal{L}_t = \log p(y, \theta[1]) - \log q(\theta[1] | \lambda_{t-1})$ \;
      $r = 1 + M (\mathcal{L}_t - \mathcal{L}_{t - 1})/\mathcal{L}_{t-1}$ \;
      \eIf{$u \leq \min(1, r)$}
        {
            $\lambda_t = \lambda_{t-1} + \rho_t \hat\nabla_\lambda \mathcal{L}(\lambda_{t-1})$ \;
            $\nu_t = 0$ \;
        }{
            $\lambda_t = \lambda_{t-1}$\;
            $\nu_t = \nu_{t - 1} + 1$ \;
        }
    }
\end{algorithm}

\subsection{YOASOVI As Metropolis Sampling}

As hinted in the discussion, the above method is naive and formulated only with the intuition that we wish $\mathcal{L}_{t}$ to be greater than or equal to $\mathcal{L}_{t-1}$ at each iteration. We briefly outline a more formal approach through the use of the Metropolis algorithm. Nevertheless, we note that this Metropolis-type algorithm is approximated by our Naive formulation, and for specific purposes that will be revealed in this discussion, we will choose to apply the naive form rather than the Metropolis form for certain desirable properties. This approach has another benefit aside from being theoretically cleaner: it gives us a clue as to how we might treat the hyperparameter $M$ which we will explore with more specific details in the next section.

\begin{figure}[ht]
  \centering
  \begin{multicols}{2}
    \includegraphics[width=\linewidth]{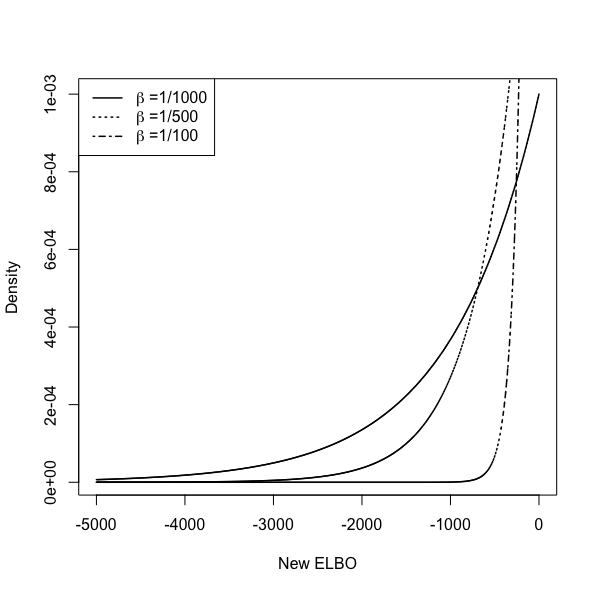}\par 
    \includegraphics[width=\linewidth]{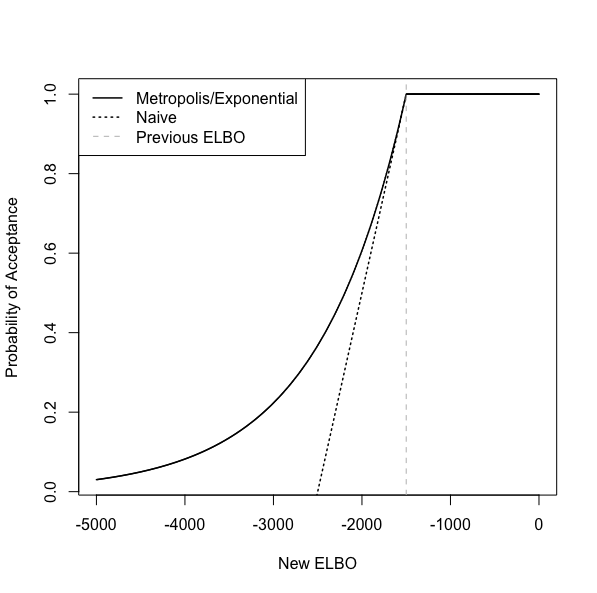}\par 
    \end{multicols}
  \caption{\label{fig:probabilities} \textit{Left}, probabilities of New ELBO values in the negative exponential distribution. The Negative Exponential allows us to represent our desired property that the ELBO should be higher rather than lower, with $\beta$ encoding how strict we are with this requirement. \textit{Right}, resulting acceptance probabilities in the Metropolis and Naive formulations of YOASOVI. Note that the naive formulation results in narrower probabilities for the same value of $M$, in this case $M = 1.5$.}
\end{figure}

Suppose that at each iteration, the ELBO $\mathcal{L}_{t}$ follows a distribution $p(\mathcal{L}_{t})$, given by the Negative Exponential distribution with mean parameter $\beta$,
\begin{align}
    p(\mathcal{L}_{t}) = \beta \exp\{\beta \mathcal{L}_{t} \}
\end{align}
The left-hand panel of Figure \ref{fig:probabilities} visualizes this distribution alongside the usual values of the ELBO $\mathcal{L}_{t}$. The Negative Exponential distribution is an explicit representation of our belief that the ELBO should be higher, rather than lower. This is in line with the stochastic optimization that is being performed, in which the ELBO should be moving upwards in the direction of zero with each spent iteration. From this distribution, the Metropolis algorithm would then perform acceptance sampling with probability $p_{accept} = \min(1, r)$ on the ratio $r$ given by
\begin{align}
    \frac{p(\mathcal{L}_{t})}{p(\mathcal{L}_{t-1})} = \exp\bigg\{ \frac{M (\mathcal{L}_{t} - \mathcal{L}_{t-1})}{\mathcal{L}_{t-1}} \bigg\}
\end{align}
after setting $\beta = M/\mathcal{L}_{t-1}$. As mentioned, this can be approximated using our Naive formulation of YOASOVI. Via a first-order Taylor approximation of this probability, we find that
\begin{align}
    \exp\bigg\{\frac{M (\mathcal{L}_{t} - \mathcal{L}_{t-1})}{\mathcal{L}_{t-1}} \bigg\} & = 1 + \frac{M (\mathcal{L}_{t} - \mathcal{L}_{t-1})}{\mathcal{L}_{t-1}} + o\bigg(\frac{M (\mathcal{L}_{t} - \mathcal{L}_{t-1})}{\mathcal{L}_{t-1}}\bigg) \\
    & \approx 1 + \frac{M (\mathcal{L}_{t} - \mathcal{L}_{t-1})}{\mathcal{L}_{t-1}}
\end{align}
giving us our naive YOASOVI acceptance sampler.

On the right-hand plot of Figure \ref{fig:probabilities} we see the resulting acceptance probabilities supposing that the previous iteration ELBO was at $\mathcal{L}_{t-1} = -1500$. Any ELBO drawn whose values are higher than this are accepted with certainty. On the other hand, the probabilities tail off with varying degrees depending on the formulation the further left we go. This controls the U-Turn behavior that we observe in both MCVI and QMCVI. Between the two formulations, however, we prefer the naive version over the Metropolis formalization as the former tails off the probability much faster despite being parametrized with the same value $M = 1.5$.

\subsection{Tempered Acceptance Sampling}

After having formalized our discussion of how the YOASOVI acceptance sampler is formulated, we can now present an intuition for the parameter $M$ that appears in both the Naive and the Metropolis formulations. We know from the negative exponential distribution that $\text{Var}(\mathcal{L}_{t}) = -\mathcal{L}_{t-1}/M$. If we set $M = 1$ then we generally allow $\mathcal{L}_{t}$ to vary by as much as $\sqrt{-\mathcal{L}_{t-1}}$. On the other hand, $M = 10$ means that we only allow $\mathcal{L}_{t}$ to vary only by as much as $\sqrt{-\mathcal{L}_{t-1}/10}$.

Performing YOASOVI could thus be done in two ways. The analyst can begin with a warm-up stage consisting of a few iterations, finding a value of $M$ that yields acceptable changes in the resulting ELBO. If the ELBO makes large U-turns, then one can simply increment $M$ upwards until this behavior has been comfortably controlled.

As an alternative, one can also set $M$ to be a function of the number of iterations, such that the variation of the resulting ELBOs are set progressively tighter into the latter iterations. For instance, setting $M(t) = k \log(t)$ for some $k > 0$ and $t = 1, 2, ...$ or $M(t) = t$ monotonically decreases the allowed variation in per-iteration ELBO, resulting in a \textit{tempered} acceptance sample. This is coming from the fact that models of the form $\exp\{-\beta E(x)\}$ are also known in the statistical physics literature as \textit{Energy Based Models} (EBM) \cite{Lecun2006}, where $\beta$ is inversely proportional to the \textit{temperature} of the system.

We can think of the acceptance sampling probability in YOASOVI as defining an energy function that assigns a score to the resulting ELBO of the Monte Carlo sample. The probability of a configuration is then defined in terms of its energy relative to the ELBO of the previous iteration. Lower energy configurations (values closer to $\mathcal{L}_{t - 1}$) are assigned higher probabilities. In this system, the temperature parameter scales the energy values before computing the probabilities. When the temperature is high, it means that the energy differences between configurations have less influence on the probabilities \cite{Arbel2021}. In other words, configurations with higher energy (i.e., wider deviations from $\mathcal{L}_{t - 1}$) are more likely to be sampled, leading to a more exploratory sampling behavior. This can be useful for escaping local energy minima and exploring the space of possible configurations more broadly.

Conversely, when the temperature is low, the energy differences between configurations have a stronger influence on the probabilities. Configurations with lower energy are more likely to be sampled, leading to a stricter sampling behavior. This can be useful for focusing on high-probability regions of the configuration space and generating more useful samples. Adjusting $M$ thus results in an algorithm that starts with high temperature to allow wider exploration of the parameter space even if they result in some U-Turns of the objective, then gradually decreasing the temperature to force the algorithm to focus on parameter updates that lead to actual improvements. This continues until no such improvements can be found, at which point the algorithm stops.

This, combined with the Early Stopping heuristic discussed in the subsequent section, ensures that the algorithm will eventually converge. As a specific case, suppose that $M(t) = k \log(t)$. We note that from the exponential distribution, the probability that any new ELBO $\mathcal{L}_t > \mathcal{L}_{t - 1}$ is given by
\begin{align}
    \lim_{t \to \infty} P(\mathcal{L}_t > \mathcal{L}_{t - 1}) & = \lim_{t \to \infty} \exp\{- k \log(t) \mathcal{L}_{t} \} \\
    & = \lim_{t \to \infty} \bigg( \frac{1}{t^k} \times \exp\{ \mathcal{L}_{t-1} \} \bigg) \\
    &= 0
\end{align}
supposing only that $\lim_{t \to \infty} \mathcal{L}_{t-1} = \mathcal{L}'$ not necessarily the maximum.

Another way of interpreting adaptive values of $M$ is through our observation that it directly controls the allowed variation for $\mathcal{L}_{t}$ around the previous value. Thus, setting $M$ to increase monotonically across iterations is the same as enforcing a prior belief that the ELBO should not be varying very widely towards the latter iterations. This means that the algorithm is free to explore a relatively wider range in the beginning of its runs, but becomes much stricter as the iterations are spent. Setting $M(t) = k \log(t)$ is simply tantamount to controlling for this strictness to not grow too quickly.

\subsection{Early Stopping}

YOASOVI also solves an ongoing problem for Stochastic VI regarding optimal convergence criteria. In the original formulation, convergence is tested for based on the magnitude of the difference between the previous variational parameter values $\lambda_{t - 1}$ and its update $\lambda_t$. A key issue with this convergence rule is that it has no direct relationship with the objective function, and can therefore cause the algorithm to stop on account of the update being small, even when it hasn't yet reached a satisfactory solution.

When performing YOASOVI, the probability of acceptance is expected to decrease monotonically as the algorithm approaches the optimal value of the ELBO, and no further improvements can be made. Thus, we can adopt the early stopping heuristic used for training deep neural networks \cite{Goodfellow-2016}. At each iteration $t$, we simply accumulate a counter $\nu_t$,
\begin{align}
    \nu_t = \begin{cases}
        \nu_{t - 1} + 1 & \hat\nabla_{\lambda} \mathcal{L}(\lambda_{t - 1}) \text{ rejected} \\
        0 & \hat\nabla_{\lambda} \mathcal{L}(\lambda_{t - 1}) \text{ accepted}
    \end{cases}
\end{align}
which will continue to increase so long as no new sample is accepted. Otherwise, $\nu_t$ resets to zero when a sample is finally accepted. Thus, setting a hyperparameter for \texttt{patience}, the algorithm is forced to stop whenever $\nu_t > \text{patience}$. For example, suppose that \texttt{patience} $= 10$,  the algorithm concludes convergence when no new sample has been accepted in the last 10 iterations.

\begin{algorithm}[ht]
    \label{alg:metropolis}
    \caption{YOASOVI with Metropolis acceptance sampling}

    \DontPrintSemicolon
    \SetAlgoLined
    \SetKwInOut{Input}{Input}\SetKwInOut{Output}{Output}
    \Input{Model, \texttt{patience} parameter, learning rate $\rho_t$}
    Initialize $\lambda_0$ randomly, set $t=0$, $\nu_t = 0$, and $\mathcal{L}_0 = -\infty$\;
    \BlankLine
    \While{$\nu_t <$ \texttt{patience}}{
      $t = t+1$\;
      // Draw a single MC sample from $q(\theta | \lambda_{t-1})$ \;
      $\theta[1] \sim q(\theta | \lambda_{t-1})$ \;
      $u \sim \text{Uniform}(0,1)$ \;
      $\hat\nabla_\lambda \mathcal{L}(\lambda_{t-1}) = \nabla_{\lambda} \log q(\theta[1] | \lambda_{t-1}) (\log p(y, \theta[1]) - \log q(\theta[1] | \lambda_{t-1}))$ \;
      $\mathcal{L}_t = \log p(y, \theta[1]) - \log q(\theta[1] | \lambda_{t-1})$ \;
      $r = \exp\{ M (\mathcal{L}_t - \mathcal{L}_{t - 1})/\mathcal{L}_{t-1} \}$ \;
      \eIf{$u \leq \min(1, r)$}
        {
            $\lambda_t = \lambda_{t-1} + \rho_t \hat\nabla_\lambda \mathcal{L}(\lambda_{t-1})$ \;
            $\nu_t = 0$ \;
        }{
            $\lambda_t = \lambda_{t-1}$\;
            $\nu_t = \nu_{t - 1} + 1$ \;
        }
    }
\end{algorithm}

\subsection{Adaptive Optimizers}

To be sure, YOASOVI is still a stochastic optimization algorithm, albeit one that uses only single-draw samples. The update step (\ref{eq:stochOpt}) is retained, for which the learning rule $\rho_t$ continues to be a consideration. In line with recent findings in both VI and the more established literature surrounding stochastic optimization, we can set $\rho_t$ to follow any adaptive learning rule such as AdaGrad \cite{Duchi-2011}, AdaDelta \cite{Zeiler-2012}, Adam \cite{Kingma-2013}, or RMSProp \cite{Goodfellow-2016}. Because under acceptance sampling, the single-sample gradient estimate is still a Monte Carlo average, the same computational gains are expected with the use of these optimizers as in QMC, MCVI, or regularized MCVI. Nevertheless, this requires extra training to find the best hyperparameters of the optimizers, which may be different for each algorithm. For the sake of simplicity, the experiments presented in Section \ref{sec:experiments} make use of only a constant learning rate, $\rho_t = 0.001$.

\section{Empirical Results}
\label{sec:experiments}

\paragraph{Simulated Datasets} We first demonstrate the performance of YOASOVI in a set of simulated Gaussian mixture distributions. For these experiments, we generate samples of size $N = 500$ from a multivariate Gaussian distribution with $K = 2, 3$ clusters in the bivariate case, and $K = 4$ in the 3-variate case. This is to stress-test the algorithms across more and more complex models. As baseline, we compare with the QMCVI implementation outlined by \citet{Buccholz-2018} using $S = 10$ Sobol sequence samples, as well as MCVI with $S = 100$ samples regularized with the positive-part James-Stein estimator (BBVI-JS+) \cite{Dayta-2024}.

\begin{table}[ht]
\caption{Convergence time, ending ELBO and DIC for BBVI-JS+, QMCVI, and YOASOVI in simulated $p$-variate Gaussian mixture distributions with $K$ clusters ($N = 500$). YOASOVI consistently finishes within 1 to 2 seconds of runtime, while QMCVI takes up to a minute. Despite this short runtime and using only 1 sample per iteration, it consistently achieves higher ELBO values and lower DIC, both indicating better fit.}
\label{tab:simulated}
\centering\begingroup\fontsize{9}{11}\selectfont

\begin{tabular}{rrlrrrrrrrr}
\toprule
\multicolumn{1}{c}{} & \multicolumn{1}{c}{} & \multicolumn{1}{c}{} & \multicolumn{2}{c}{Iterations} & \multicolumn{2}{c}{Time (secs)} & \multicolumn{2}{c}{ELBO} & \multicolumn{2}{c}{DIC} \\
\cmidrule(l{3pt}r{3pt}){4-5} \cmidrule(l{3pt}r{3pt}){6-7} \cmidrule(l{3pt}r{3pt}){8-9} \cmidrule(l{3pt}r{3pt}){10-11}
$p$ & $K$ & Method & Mean & SD & Mean & SD & Mean & SD & Mean & SD\\
\midrule
2 & 2 & BBVI-JS+(100) & 500.0 & 0.0 & 137.0 & 60.3 & -4,190.0 & 553.8 & 8,058.2 & 1098.9\\
  &   & QMCVI(10) & 500.0 & 0.0 & 31.7 & 15.6 & -5,104.1 & 0.0 & 9,615.5 & 0.0\\
  &   & YOASOVI(1) & 175.5 & 40.8 & 0.6 & 0.3 & -4,005.4 & 224.8 & 8,015.6 & 446.6\\
\addlinespace
2 & 3 & BBVI-JS+(100) & 251.2 & 262.3 & 51.2 & 55.3 & -4,621.7 & 767.3 & 9,146.6 & 1606.6\\
  &   & QMCVI(10) & 500.0 & 0.0 & 45.2 & 0.8 & -6,501.5 & 0.0 & 12,160.3 & 0.0\\
  &   & YOASOVI(1) & 350.1 & 160.3 & 1.5 & 0.7 & -3,713.4 & 135.5 & 7,427.0 & 269.0\\
\addlinespace
3 & 4 & BBVI-JS+(100) & 53.0 & 157.1 & 3.1 & 7.8 & -5,348.8 & 570.1 & 10,636.1 & 1206.3\\
  &   & QMCVI(10) & 500.0 & 0.0 & 39.0 & 14.0 & -6,563.3 & 0.0 & 12,160.3 & 0.0\\
  &   & YOASOVI(1) & 291.4 & 198.2 & 1.1 & 0.7 & -3,950.1 & 206.0 & 7,900.8 & 407.9\\
\bottomrule
\end{tabular}
\endgroup{}
\end{table}

The results are summarized in Table \ref{tab:simulated}. In the bivariate case with $K = 2$ clusters, YOASOVI is able to complete within $0.6 \mp 0.3$ seconds, whereas QMCVI take about $31.7 \mp 15.6$ seconds. However, this time should be measured as more of a lower-bound, since QMCVI was not able to reach convergence within 500 iterations, at which point the runs were halted. Similarly, BBVI-JS+ also did not converge within 500 iterations despite making use of $S = 100$ samples.

This result is generally consistent across all runs: YOASOVI converges within 1 to 2 minutes of runtime. Despite this short run-time and making use of a single sample, the tempered acceptance sampling procedure has controlled the trajectory of its ELBO such that it ends up within neighborhoods of much higher ELBO, and lower Deviance Information Criterion (DIC) \cite{Gelman-2013}, both indicating better fit than where BBVI-JS+ and QMCVI have ended up by the end of their runs (both averaging at 1-2 minutes).

\begin{figure}[ht]
  \centering
  \includegraphics[width=\linewidth]{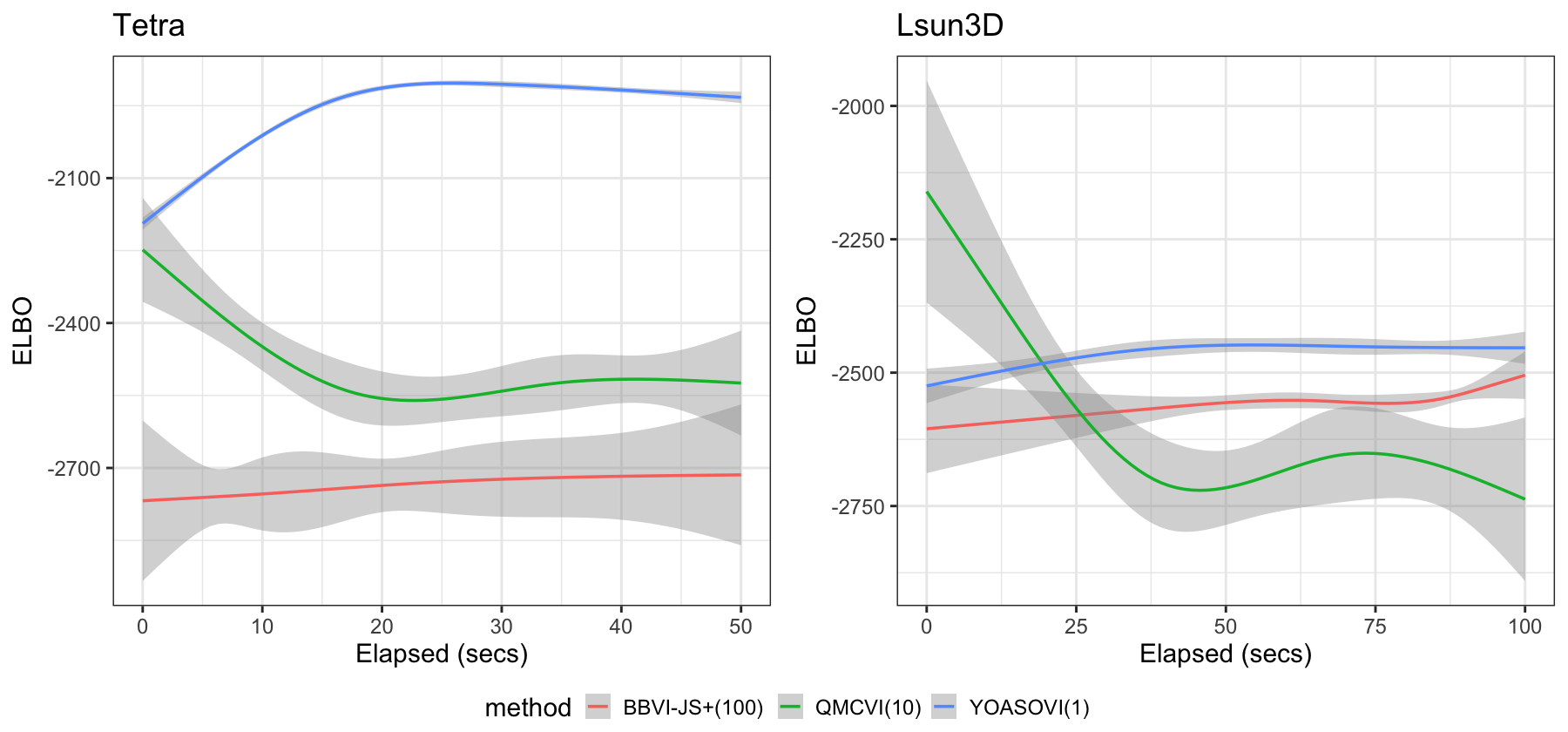}
  \caption{\label{fig:benchmarks} ELBO trajectories of BBVI-JS+, QMCVI, and YOASOVI within the first 50 and 100 seconds of the algorithm in two benchmark datasets from \cite{FCPS}. YOASOVI separates itself well from the other two algorithms with a strongly increasing trajectory.}
\end{figure}

\paragraph{Benchmark Datasets. } We also apply the algorithm to some fundamental clustering benchmark datasets provided by \citet{FCPS} in the \texttt{FCPS} package for \texttt{R}. In Figure \ref{fig:benchmarks} we plot the trajectories of the algorithms within 50 and 100 seconds of their run. It should be noted that for both datasets, YOASOVI has generally already converged within this range, while QMCVI is still warming up, while BBVI-JS+ has only just started its run. Nevertheless, the performance of YOASOVI is clearly well-separated from the other two algorithms with a strongly increasing trajectory for its ELBO.

\section{Conclusions}
\label{sec:conclusions}

In this paper, we have proposed YOASOVI, an algorithm for fast, self-correcting Monte Carlo Variational Inference through an acceptance sampling scheme. To achieve this, we take advantage of the availability of the ELBO objective function value at each iteration and accept or reject samples based on whether the resulting ELBO is indeed an improvement over the previous iteration. Incorporating some concepts from statistical physics in the form of tempered energy-based models, we are able to create a tempered version of the algorithm that becomes increasingly strict at improving the ELBO towards the latter runs, ensuring the algorithm does converge within an optimal neighborhood of the objective function.

Experiments involving both simulated and benchmark datasets in Gaussian mixture models show consistent strong performance by YOASOVI at beating both regularized MCVI and Quasi-Monte Carlo VI in model fit and time to convergence. Therefore, YOASOVI provides a promising framework for performing MCVI that is both fast and reliable, without overburdening the algorithm with layers of architectural modifications and calibrations, and retaining the generality of the method. Its inexpensive computation and generality means that it should be straightforward to apply to any existing software environment currently being used for machine learning applications, such as R or Python, and can be run reliably even on underpowered personal machines.

\section*{References}
\bibliographystyle{unsrtnat}

{
    \small
    \bibliography{mc_bbvi}
}

\end{document}